# COVID-19 Electrocardiograms Classification using CNN Models

Ismail Shahin
Department of Electrical Eng., University of Sharjah, Sharjah, United Arab Emirates
ismail@sharjah.ac.ae

Ali Bou Nassif
Department of Computer Eng., University of Sharjah, Sharjah, United Arab Emirates
anassif@sharjah.ac.ae

Mohamed Bader Alsabek
Department of Electrical Eng., University of Sharjah, Sharjah, United Arab Emirates
u16104773@sharjah.ac.ae

*Abstract*— With the periodic rise and fall of COVID-19 and numerous countries being affected by its ramifications, there has been a tremendous amount of work that has been done by scientists, researchers, and doctors all over the world. Prompt intervention is keenly needed to tackle the unconscionable dissemination of the disease. The implementation of Artificial Intelligence (AI) has made a significant contribution to the digital health district by applying the fundamentals of deep learning algorithms. In this study, a novel approach is proposed to automatically diagnose the COVID-19 by the utilization of Electrocardiogram (ECG) data with the integration of deep learning algorithms, specifically the Convolutional Neural Network (CNN) models. Several CNN models have been utilized in this proposed framework, including VGG16, VGG19, InceptionResnetv2, InceptionV3, Resnet50, and Densenet201. The VGG16 model has outperformed the rest of the models, with an accuracy of 85.92%. Our results show a relatively low accuracy in the rest of the models compared to the VGG16 model, which is due to the small size of the utilized dataset, in addition to the exclusive utilization of the Grid search hyperparameters optimization approach for the VGG16 model only. Moreover, our results are preparatory, and there is a possibility to enhance the accuracy of all models by further expanding the dataset and adapting a suitable hyperparameters optimization technique.

## I. Introduction

Since the horrific outbreak of COVID-19 in March 2020, the virus has widely disseminated among the globe, setting a record of more than 242 million confirmed cases and an increasing number of fatalities, surpassing 4 million confirmed deaths worldwide as per the World Health Organization (WHO) statistics [1]. It has been observed that the most common and specific symptoms of this virus are fever and cough, with some other non-specific symptoms such as fatigue, headache, and dyspnea [2]. The virus has attracted an intimidating concern, due to its unbelievable spread among people, in addition to its critical damages that have targeted the respiratory system [3]. Moreover, it has been observed that it has a vital effect on the cardiovascular system, resulting in multi-organ failure [4]. Thus, the Electrocardiograms (ECG) analysis can provide a fast and cost-efficient technique to detect the presence of COVID-19 infection. ECG is a graph of voltage versus time of the heart's electrical activity that is recorded by the electrodes placed on the skin [5]. The preposterous dissemination of this deadly virus has led the health industries to keenly look for new technologies to be employed against the ongoing COVID-19 crisis. Lately, deep learning techniques have been immensely utilized in detecting, classifying, and preventing COVID-19 infection [6].

The rest of the paper is organized as follows: Section II covers the related work. Section III demonstrates the methodology of the proposed model. Section IV provides the experiments and results. While Section V discusses the concluding remarks.

## II. Literature Review

Artificial Intelligence (AI) has been heavily utilized in the digital health sector to provide aid in the process of early COVID-19 detection [7]. Researchers around the world have shown that the detection of COVID-19 by the analysis of respiratory sounds, chest X-rays, CT scans, and the ECG is feasible by the implementation of AI, specifically the deep learning techniques [8]. The utilization of the respiratory data has shown remarkable results in the COVID-19 detection. In [9], authors have proposed a deep learning-based system named CovNet, that can perform analysis and classifications on the cough sound recordings of both positive and negative COVID-19 acoustic samples. The Mel Frequency Cepstral Coefficients (MFCCs) have been utilized as the feature input for the proposed model. The system has achieved an Area Under Curve (AUC) score percentage of 72.23% before data augmentation and 87.07% after the utilization of data augmentation. Besides, Hassan *et al.* [3] have proposed a speech-based deep learning model that detects the presence of COVID-19 from the cough, breathing, and voice acoustics related to both COVID-19 and non-COVID-19 samples, The Long Short-Term Memory (LSTM) has been employed for the classification. The proposed system has achieved an accuracy of 88.2% for the voice test, which is relatively low compared to the cough and breathing sound samples, which have achieved accuracies of 98.2% and 97%, respectively. A variety of studies have utilized deep learning for the COVID-19 identification, which is mainly based on CT scans and X-ray images. Nevertheless, Aside from the high prevalence of COVID-19 diagnosis success, the analysis of electrocardiograms has been found very useful in the aiding process of early COVID-19 detection.

in [10], authors have proposed a novel method to diagnose COVID-19 using deep learning Techniques on the ECG dataset. Additionally, authors have used a modern, effective method to represent 12-lead ECG to 2-D colorful images. Moreover, the features have been extracted using the Gray-level co-occurrence matrix (GLCM). Furthermore, the obtained generated images are fed into the designed CNN for the COVID-19 diagnosing. Two classification methods were conducted for the COVID-19 classification. In the first one, the combination of ECG data from both positive COVID-19 and no-findings (normal) has been classified, obtaining an accuracy of 96.20%, while, in the second method, the ECG combination that is comprised of negative (normal, abnormal, and myocardial-infraction) and positive (COVID-19) have been evaluated, obtaining an accuracy of 93%. Tawsifur *et al.* [4] have proposed a deep CNN-based system for the detection of COVID-19 based on the ECG trace images. The utilized public data set consists of 1937 images that are comprised of five separate categories: normal, COVID-19, myocardial infarction, abnormal heartbeat, and the recovered myocardial infraction. Furthermore, in this study, 6 pre-trained convolutional network architectures have been utilized for the classification phase, which are DensNet201, Inceptionv3, ResNet18, ResNet50, ResNet101, and MobileNetv2 to classify three classification schemes which are: two-class classification (Normal vs. COVID-19), three class classification (Normal vs. COVID-19 vs. CVDS), five class classification (Normal vs. COVID-19 vs. MI vs. AHB and RMI). The DenseNet201 model is superior to the other models on classifying the first two classes, obtaining an accuracy of 99.1% and 97.36%, respectively. Besides, the Inceptionv3 model has also outperformed the remaining classifiers for the last class (five classes) classification, yielding an accuracy of 97.83%.

III. METHODOLOGY

*A. Dataset*

In this proposed framework, data collection is an essential part of initiating the training of the deep learning algorithm. The dataset has been gathered from the dataset that contains ECG images of cardiac and COVID-19 patients. The dataset was shared online by Khan et al. [11]. The dataset comprises of 1109 individuals. A 12-lead system has been utilized to obtain the ECG drawings from the acquired dataset.

*B. Image Pre-processing*

Image pre-processing has been implemented to eliminate undesirable distortions. Unfortunately, the images suffer from insufficient resolution. In addition, the sizes of the images are not standard. The original image, as shown in Fig. 1, has been subjected to a simple and effective segmentation approach to avoid the high computational costs and information loss that occurs in the complex image pre-processing. In the proposed technique, firstly, the electrocardiogram images have been segmented, the process of segmentation has been executed within a rectangular frame. Due to the fact that the resolutions of the electrocardiogram images in the dataset vary. Secondly, a density map function has been employed for the filtration of the input densities to eliminate the paper lines of the ECG images. Fig. 2, provides an illustration for a segmented ECG image without the lines of the background. Additionally, scaling and resizing of the images have been made to provide the exact size of images to the models. The original size of the images was 2213 × 1572 pixels. For the experimental setup, the size of the images has been scaled to 987 × 987 pixels. Furthermore, deep neural networks demand a relatively large corpus to learn efficiently. Thus, the data augmentation technique has been utilized to artificially inflate the dataset by applying several approaches and noises to the original images, such as modifying the brightness, zooming, horizontal or vertical mirroring, etc. The previously mentioned distortions must not alter the spatial pattern of the target classes [12], [13].

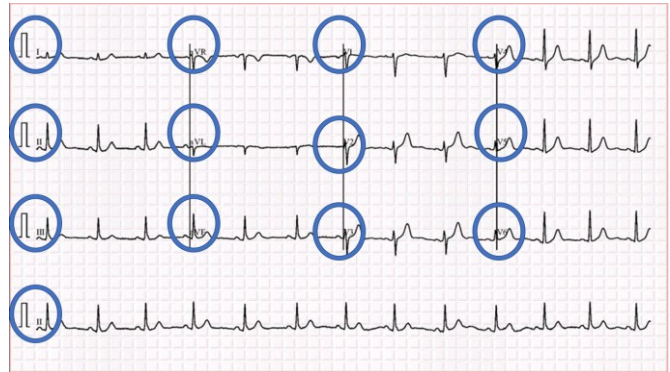

Figure 1: The acquired ECG signal image

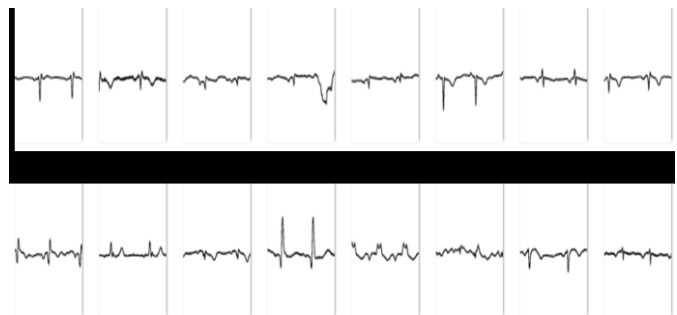

Figure 2: The filtered image

*C. Convolutional Neural Network (CNN)*

Convolutional Neural Network (CNN) is a subset of deep learning which has been significantly utilized for solving complex image-driven pattern recognition tasks [11], [14]. CNNs principally concentrate on the core that the input will be comprised of images. This concentrates the architecture to be established in the most suitable way needed for dealing with the specific type of data [15]. CNNs are composed of 3 sorts of layers. namely, convolutional layers, pooling layers, and fully connected layers. The convolutional and fully connected layers have parameters, but pooling and non-linearity layers don't have parameters [16]. Every layer is made up of a high number of interconnected computational nodes referred to as neurons. A CNN architecture is constructed when these layers are stacked.

*D. CNN pre-trained models*

Transfer learning is typically employed when there is a new dataset smaller than the original dataset utilized for the training of the pre-trained model [17]. The main benefit of transfer learning is that it reduces the time taken to develop and train a model by reusing weights of already developed models [18]. In this paper, several CNN models have been employed for image classification.

1- **ResNet50**: ResNet is a typical feed-forward network that is residually connected. The residual network comprises of various fundamental residual blocks. Nevertheless, the residual block operations varies depending on the diverse architecture of residual networks [18], [19]. The ResNet50 is a short form of residual network that has 50 layers.

2- **DenseNet-201**: DenseNet201 architecture is inspired by the building block of ResNet. It was introduced to deal with the vanishing-gradient problem by introducing a simple connectivity pattern between layers. It is comprised of multiple layers that are densely connected, where the outputs of the layers are connected to all the successors in what is called the dense block [19].

3- **VGG16 and VGG19:** VGG Net is one of the most reliable CNN models. Its unique architecture contains an approximation of 138 to 144 million parameters, including 16 or 19 convolutional layers for VGG16 and VGG19, respectively [18], [19].

4- **Inceptionv3:** Inceptionv3 is commonly utilized to enhance the computing resources by the increment of the depth and width of the network [20]; the model is comprised of 48 layers.

5- **Inceptionresnetv2:** Represents a combination of inception structure combined with residual connections that include 164 deep layers [20].

*E. Fine-tuning*

Fine-tuning aims to store the gained knowledge while solving one dilemma and employing it to a distinct dilemma [21]. Fine-tuning is a remarkable way to enhance the performance of the system by fine-tuning the weights of the pre-trained model top layers. There are two methods for fine-tuning; one is to freeze the weight of particular layers. The more frozen layers there are, the less fine-tuning effort is needed regarding time and resources. The other approach would be to train the complete architecture on a new database. [22].

*F. Grid Search*

Grid Search (GS) is one of the most popular algorithms for hyperparameters optimization. Basically, in GS, the range of possible parameters is set manually. Afterward, the program starts a complete search over these subsets of the hyperparameters space of the training algorithm. Grid search goes through all the possible combinations based on the provided values [23]. In this algorithm, all possible hyperparameters combinations are being brute-forced, then the models are evaluated using the cross-validation technique [24]. In this proposed system, the grid search technique has been only utilized for the VGG16 CNN model for the selection of the most optimum hyperparameters of batch size, dropout, epochs, and learning rate, which are 32, 0.1, 25, and 0.001, respectively.

IV. EXPERIMENTAL RESULTS

In the proposed CNN model, the grid search optimization technique has only been utilized for the VGG16 model to obtain the most optimum hyperparameters of batch size, dropout, epochs, and learning rate, which are 32, 0.1, 25, and 0.001, respectively. Table I illustrates the grid search results for different values of hyperparameters, which were set for the VGG16 model. Moreover, data augmentation has been utilized to enhance the accuracy of the system by artificially inflating the dataset. Besides, the K-fold cross-validation technique has been employed to overcome the generalization problems. Fig. 3 demonstrates the methodology that has been followed in implementing the COVID-19 detection system based on ECG images. For this system, a binary classification has been conducted to detect the presence of COVID-19. The split of the utilized dataset has been done as follows: 70% of the dataset has been utilized for training, 20% for the testing, and 10% for the validation. Two tests were made to evaluate the performance of the system. In the first test, six different CNN models were used, namely, VGG16, VGG19, ResNet50, DenseNet201, InceptionV2, and InceptionResNetV2. The models have been employed as feature extractors in the system without any fine-tuning. As shown in Fig. 4, the VGG16 model has outperformed all the models, obtaining the highest accuracy of 81.39%. Since the utilized data is relatively small, the architecture of the VGG16 model is considered to be more suitable than the rest of the proposed models. Furthermore, to improve the system's accuracy, the VGG16 model has been fine-tuned in the second test. Thus, it has achieved the highest performance. As a result of the fine-tuning, the system has obtained an accuracy of 85.92%; also, the loss of the system has been decreased.

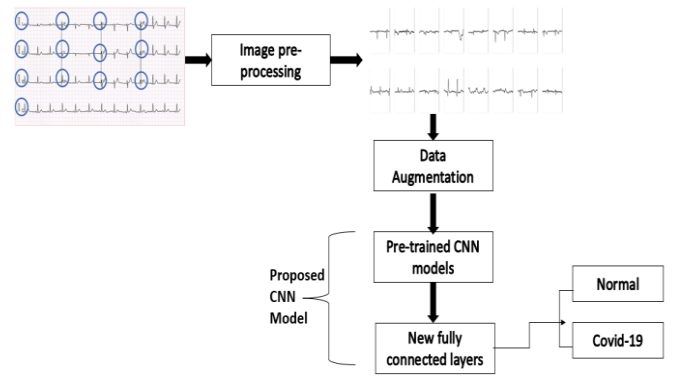

Figure 3: Methodology for detecting COVID-19 from ECG signals

This paper provides an innovative deep Convolutional Neural Network-based transfer learning approach for early COVID-19 diagnosis. In the proposed framework, several techniques have been employed to enhance the performance of the system, such as transfer learning, grid search optimization, fine-tuning, and data augmentation. The VGG16 CNN pre-trained model has outperformed all the other models by obtaining an accuracy of 81.39%. Furthermore, fine-tuning techniques have been done on VGG16. Thus, the accuracy of the system has been enhanced to reach 89.64%, outperforming all of the other proposed models. The grid search technique has only been utilized for the VGG16 model, which is due to the variety of combinations it provides. The reason behind not utilizing it for the rest of the models is that it will require a huge amount of time to obtain the most optimum hyperparameters for the rest of the models.

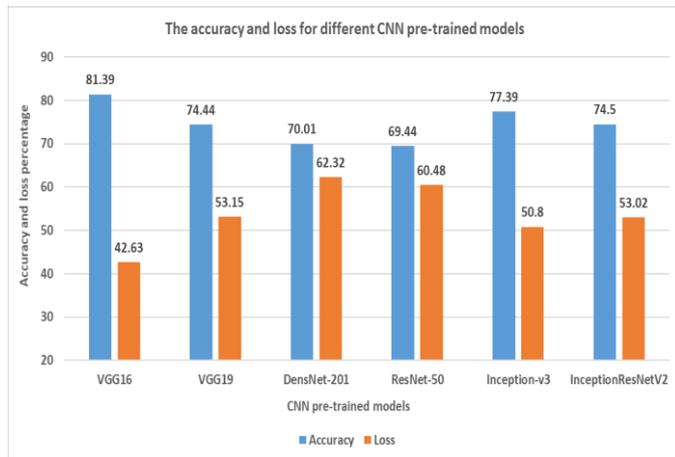

Figure 4: The accuracy and loss of different CNN pre-trained models

TABLE I. MODEL PERFORMANCE FOR DIFFERENT VALUES OF HYPERPARAMETERS

| Epoch | Batch Size | Dropout | Layer Neurons | Accuracy | Error |
|---|---|---|---|---|---|
| 25 | 16 | 0.1 | 16 | 50 | 0.6934 |
| | | | 32 | 0.6983 | 0.5379 |
| | | | 64 | 0.7967 | 0.4557 |
| | | 0.2 | 16 | 0.7761 | 0.4922 |
| | | | 32 | 0.7228 | 0.5510 |
| | | | 64 | 0.7983 | 0.4553 |
| | | 0.5 | 16 | 0.7800 | 0.4982 |
| | | | 32 | 0.7878 | 0.4845 |
| | | | 64 | 0.5728 | 0.9073 |
| | **32** | **0.1** | 16 | 0.7894 | 0.4546 |
| | | | **32** | **0.8139** | **0.4263** |
| | | | 64 | 0.7911 | 0.4455 |
| | | 0.2 | 16 | 0.7467 | 0.4961 |
| | | | 32 | 0.5000 | 0.6934 |
| | | | 64 | 0.7389 | 0.4994 |
| | | 0.5 | 16 | 0.6934 | 0.5000 |
| | | | 32 | 0.7756 | 0.4815 |
| | | | 64 | 0.7894 | 0.4708 |
| | 64 | 0.1 | 16 | 0.5000 | 0.6932 |
| | | | 32 | 0.7633 | 0.4729 |
| | | | 64 | 0.8133 | 0.4212 |
| | | 0.2 | 16 | 0.5000 | 0.6932 |
| | | | 32 | 0.5016 | 0.6932 |
| | | | 64 | 0.8094 | 0.4442 |
| | | 0.5 | 16 | 0.6932 | 0.5000 |
| | | | 32 | 0.5000 | 0.6938 |
| | | | 64 | 0.6922 | 0.5388 |
| 50 | 16 | 0.1 | 16 | 0.8111 | 0.4507 |
| | | | 32 | 0.8106 | 0.4291 |
| | | | 64 | 0.7950 | 0.4620 |
| | | 0.2 | 16 | 0.7422 | 0.5084 |
| | | | 32 | 0.7278 | 0.4967 |
| | | | 64 | 0.7044 | 0.5251 |
| | | 0.5 | 16 | 0.8078 | 0.4586 |
| | | | 32 | 0.5000 | 0.6933 |
| | | | 64 | 0.7783 | 0.4833 |

## V. CONCLUDING REMARKS

In this proposed system, the performance is non-ideal, which is due to the relatively small dataset and the noisiness in the acquired images. The obtained results are preliminary and subject to further enhancement. A hybrid CNN-LSTM model and several techniques such as oversampling and fine-tuning could be implemented to enhance the system's performance. Eventually, it can be concluded that the automatic detection of cardiovascular abnormalities is feasible within the framework of deep learning.

## ACKNOWLEDGMENTS

"The authors would like to thank the University of Sharjah in the United Arab Emirates for funding this work through the competitive research project entitled "Emirati-Accented Speaker and Emotion Recognition Based on Deep Neural Network, No. 19020403139".